\documentclass[10pt,twocolumn,letterpaper]{article}

\usepackage{soul}
\usepackage{array}
\usepackage{blindtext}
\usepackage{style/cvpr}
\usepackage{fancyhdr}

\usepackage{times}
\usepackage{relsize}
\usepackage[T1]{fontenc}
\usepackage[latin1]{inputenc} 
\usepackage[english]{babel}

\usepackage{epsfig}
\usepackage{graphicx}
\usepackage{wrapfig}
\usepackage{subfig}
\usepackage[belowskip=0pt,aboveskip=0pt,font=small]{caption}
\usepackage{amsmath, amsthm, amssymb}
\usepackage{textcomp}
\usepackage{stmaryrd}
\usepackage{upgreek}
\usepackage{bm}
\usepackage{cases}
\usepackage{mathtools}

\usepackage{cite}
\usepackage[pagebackref=true,breaklinks=true,letterpaper=true,colorlinks,bookmarks=false,citecolor=red]{hyperref}

\usepackage{algorithm2e}
\usepackage{algpseudocode}

\usepackage{multirow}
\usepackage{rotating}
\usepackage{booktabs}

\usepackage{enumitem}
\usepackage[olditem,oldenum]{paralist}

\usepackage{alltt}
\usepackage{listings}

\belowdisplayskip \abovedisplayskip

\newlength{\sectionReduceTop}
\newlength{\sectionReduceBot}
\newlength{\subsectionReduceTop}
\newlength{\subsectionReduceBot}
\newlength{\abstractReduceTop}
\newlength{\abstractReduceBot}
\newlength{\captionReduceTop}
\newlength{\captionReduceBot}
\newlength{\subsubsectionReduceTop}
\newlength{\subsubsectionReduceBot}

\newlength{\eqnReduceTop}
\newlength{\eqnReduceBot}

\newlength{\horSkip}
\newlength{\verSkip}

\newlength{\figureHeight}
\usepackage{style/mysymbols}

\usepackage{url}
\usepackage{xspace}
\usepackage{comment}
\usepackage{color}
\usepackage{xcolor}
\usepackage{afterpage}
\usepackage{pdfpages}
\usepackage{framed}
\usepackage{fancybox}
\usepackage{cuted}

\newcommand{\brnns}{Bidirectional RNNs\xspace}
\newcommand{\urnns}{Unidirectional RNNs\xspace}
\newcommand{\urnn}{Unidirectional RNN\xspace}

\newcommand{\topB}{\mathop{\mathrm{top\text{-}B}}}

\newcommand{\xb}{\mathbf{x}}

\newcommand{\yb}{\mathbf{y}}

\newcommand{\bb}{\mathbf{b}}
\newcommand{\hb}{\mathbf{h}}

\renewcommand{\sb}{\mathbf{s}}

\newcommand{\Yb}{\mathbf{Y}}

\cvprfinalcopy 

\def\cvprPaperID{3219} 

\ifcvprfinal\fi
\begin{document}

\pagestyle{fancy}
\renewcommand{\headrulewidth}{0pt}
\fancyhf{}
\chead{\vspace{-50pt}\large Published at IEEE Conference on Computer Vision and Pattern Recognition (CVPR) 2017\vspace{20pt}}

\title{Bidirectional Beam Search: Forward-Backward Inference in \\Neural Sequence Models for Fill-in-the-Blank Image Captioning}
\author{Qing Sun\\
Virginia Tech\\
{\tt\small sunqing@vt.edu}
\and
Stefan Lee\\
Virginia Tech\\
{\tt\small steflee@vt.edu}
\and
Dhruv Batra\\
Georgia Tech\\
{\tt\small dbatra@gatech.edu}
}

\maketitle


\vspace{\abstractReduceTop}
\begin{abstract} 
\vspace{\abstractReduceBot}
We develop the first approximate inference algorithm for 1-Best (and M-Best) decoding in bidirectional neural sequence models by extending Beam Search (BS) to reason about both forward and backward time dependencies. 

Beam Search (BS) is a widely used approximate inference algorithm for decoding sequences from unidirectional neural sequence models. Interestingly, approximate inference in bidirectional models remains an open problem, despite their significant advantage in modeling information from both the past and future. To enable the use of bidirectional models, we present Bidirectional Beam Search (BiBS), an efficient algorithm for approximate bidirectional inference.

 To evaluate our method and as an interesting problem in its own right, we introduce a novel Fill-in-the-Blank Image Captioning task which requires reasoning about both past and future sentence structure to reconstruct sensible image descriptions. 
We use this task as well as the Visual Madlibs dataset to demonstrate the effectiveness of our approach, consistently outperforming all baseline methods.
\end{abstract} 


\vspace{-10pt}
\vspace{\sectionReduceTop}
\section{Introduction}
\label{sec:intro}
\vspace{\sectionReduceBot}

\begin{figure}[t]
\centering
\includegraphics[width=0.90\columnwidth]{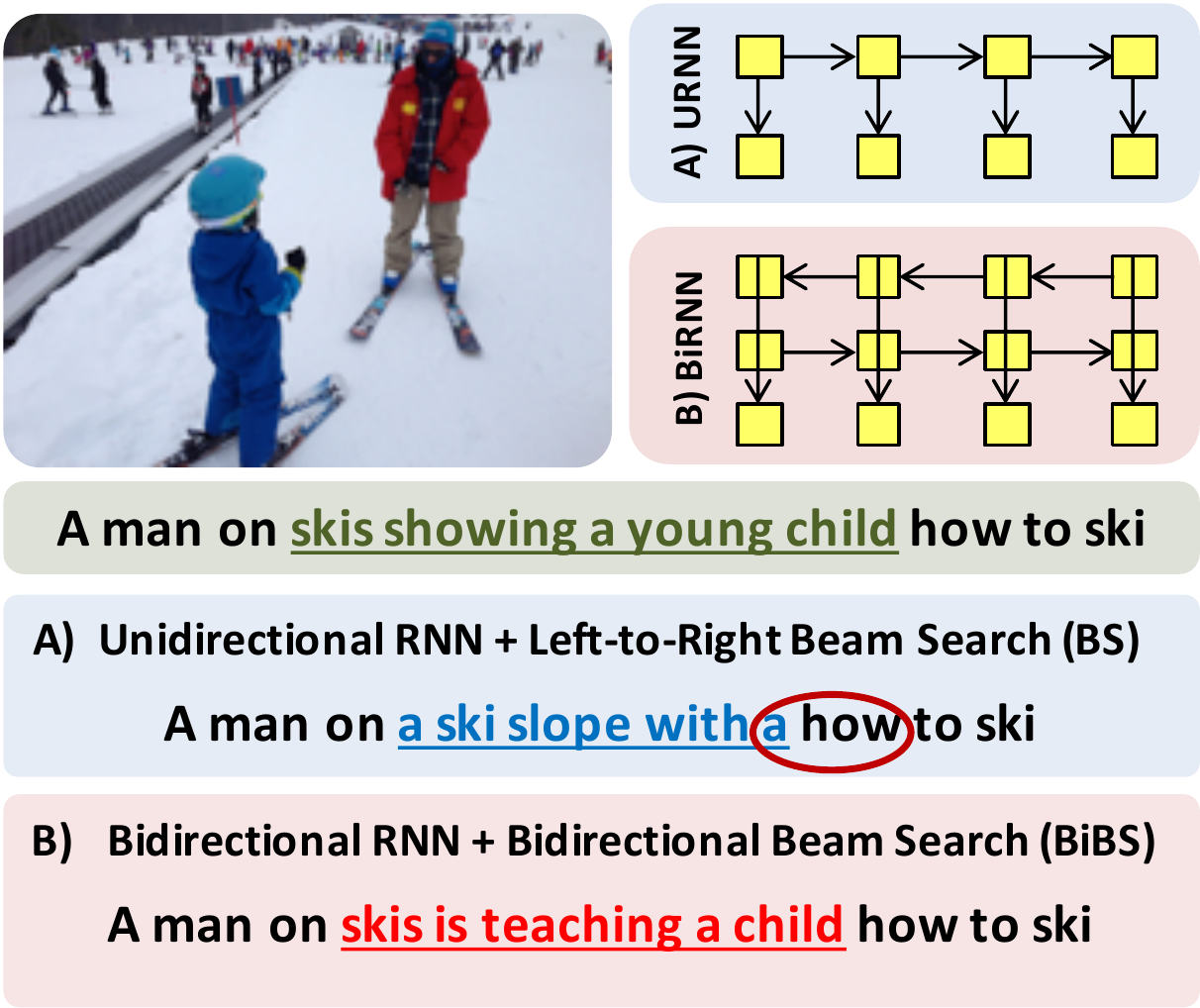}\\[10pt]
\caption{ We develop a novel Bidirectional Beam Search (BiBS) algorithm for neural sequence models and propose a new fill-in-the-blank image 
captioning task as a challenging testbed for sequence completion. Unidirectional RNNs fail to reason about both past and future outputs and 
produce nonsensical outputs for this task -- note the jarring ``\underline{with a} how to'' transition produced by classical beam search in (A). 
In contrast, our BiBS algorithm on a Bidirectional RNN produces significantly better completions (B) by considering context on either side of the blank. }
\label{fig:teaser}
\vspace{-14pt}
\end{figure}

Recurrent Neural Networks (RNNs) and their generalizations (LSTMs, GRUs, \etc) have emerged as popular and effective frameworks for modeling sequential data across varied domains.  
The application of these models has led to significantly improved performance on a variety of tasks --  speech recognition 
\cite{hinton_2012, dahl_2012}, machine translation \cite{bahdanau2014neural,cho2014learning,kalchbrenner2013recurrent}, 
conversation modeling \cite{oriol_neural_2015},  
image captioning~\cite{vinyals_cvpr15,karpathy_cvpr15,fang_cvpr15,chen_cvpr15,donahue_cvpr15}, visual question answering (VQA) \cite{antol_iccv15, ren_nips15,LuXPS16,fritz_nips14,geman_pnas14}, and visual dialog~\cite{das_cvpr16, devries_cvpr17}.

Broadly speaking, in these applications RNNs are typically used in two distinct roles -- 
(1) as \emph{encoders} that convert sequential data into real-valued vectors, and 
(2) as \emph{decoders} that convert encoded vectors into sequential output. 
Models for image caption retrieval and VQA (with classification over answers) \cite{antol_iccv15,ren_nips15} 
consist of encoder RNNs but not decoders. Image caption generation models \cite{vinyals_cvpr15} consist of decoder RNNs 
but not encoders (image encoding is performed via Convolutional Neural Networks). 
Visual dialog models use encoders to embed dialog history and model state, while using decoders to generate dialog responses.
Regardless of the setting, the task of decoding a sequence from an RNN consists of finding the most likely sequence $Y = (\yb_1, ... , \yb_T)$ given some input 
$\xb$.

Unidirectional RNNs model this probability by estimating the likelihood of outputting a symbol at time 
$t$ (say $\yb_t$) given the history of previous outputs ($\yb_1, \ldots, \yb_{t-1}$) by ``compressing'' the history into 
a hidden state vector $\hb_{t-1}$ such that $\pr(\yb_t \mid \yb_1, \ldots, \yb_{t-1}, \xb) \simeq  \pr(\yb_t \mid \hb_{t-1})$. 
Since each output symbol is conditioned on all previous outputs, the search space of possible sequences is exponential in the sequence length
and exact inference is intractable. 
As a result, approximate inference algorithm are applied, with Beam Search (BS) being the primary workhorse. 
BS is a greedy heuristic search that maintains the top-$B$ most likely partial-sequences through the search tree, where $B$ is  
referred to as the \emph{beam width}. At each time step, BS \emph{expands} these $B$ partial sequences to all possible beam extensions and 
then selects the $B$ highest scoring among the expansions. 

In contrast to Unidirectional RNNs, Bidirectional RNNs model both forward (increasing time) and backward (decreasing time) 
dependencies $\pr(\yb_t \mid \hb^f_{t}, \hb^b_{t})$ via two hidden state vectors $\hb^f_{t}$ and $\hb^b_{t}$. 
This enables \brnns to consider both past and future when predicting an output. Unfortunately, these dependencies also make 
exact inference in these models more difficult than in Unidirectional RNNs and to the best of our knowledge no efficient 
approximate algorithms exist. In this paper, we present the first efficient approximate inference algorithm for these models. 

As a challenging testbed for our method, we propose a fill-in-the-blank image captioning task. As an example, given the blanked image caption \emph{``A man on \_\_\_\_\_\_\_\_\_\_\_\_\_\_\_\_ how to ski''} shown in Figure \ref{fig:teaser}, 
our goal is to generate the missing content \emph{``\underline{skis showing a young child}''} (or an acceptable paraphrase) to complete the sentence. 
This task serves as a concrete stand-in for a broad class of 
other similar sequence completion tasks, such as predicting missing sections in a DNA sequence or path planning problems where an agent must hit intermediate checkpoints. 

On the surface, this task perhaps seems easier than generating an entire caption from scratch; there is after all, more information 
in the input. 
However, the need to condition on the context when generating the missing symbols is challenging for existing greedy 
approximate inference algorithms. Figure \ref{fig:teaser}(a) shows a sample decoding from standard `left-to-right' BS on a Unidirectional RNN. 
Note the grammatically incorrect ``\underline{with a} how to'' transition produced. 
Similar problems occur at the other boundary for right-to-left models.
Simply put, the inability to consider both the future and past contexts in BS leads Unidirectional RNNs to fill the blank with words that clash abruptly 
with the context around the blank. 

Moreover, decoding also poses a computational challenge.
Consider the following sentence that we know has only a single word missing: \emph{``The \_\_\_\_\_ was barreling down the tracks.''} 
Filling in this blank feels simple -- we just need to find the best single word $y_t^*$ in the vocabulary $\calY$. 
However, since all future outputs in a Unidirectional RNN are conditioned on the past, selecting the best word 
at time $t$ requires evaluating the likelihood \emph{of the entire sequence} once for each possible word (similarly for Bidirectional RNNs). 
This amounts to $T|\calY|$ forward passes through an RNN's basic computational unit \emph{to fill in a single blank optimally}!  
More generally, for an arbitrarily sized blank covering $w$ words, this number grows exponentially as $T|\calY|^{w}$ 
and quickly becomes intractable. 

To overcome these shortcomings, we introduce the first approximate inference algorithm for 1-Best (and M-Best) inference in bidirectional neural sequence models (RNNs, LSTMs, GRUs, \etc) -- Bidirectional Beam Search (BiBS). We show BiBS performs well on fill-in-the-blank tasks, efficiently incorporating both forward and backward time information from Bidirectional RNNs.

To give an algorithmic overview, we begin by decomposing a Bidirectional RNN into two calibrated but independent Unidirectional RNNs 
(one going forward in time and the other backward). To perform approximate inference with these decomposed models, 
our method alternatively performs BS on one direction while holding the beams in the opposite direction fixed. 
The fixed, oppositely-directed beams are used to roughly approximate the conditional probability of all future sequence given 
the past such that a BS-like update minimizes an approximation of the full joint at each time step. 
Figure \ref{fig:teaser}(b) shows an example result of our algorithm -- 
``A man on \underline{skis is teaching a child} how to ski'' -- 
which smoothly fits within its context while still describing the image content. 

We compare BiBS against natural ablations and baselines for fill-in-the-blank tasks. Our results show that BiBS is an effective and efficient approach for decoding Bidirectional 
RNNs, consistently outperforming all baselines.

\vspace{\sectionReduceTop}
\section{Related Work}
\label{sec:related}
\vspace{\sectionReduceBot}

While Unidirectional RNNs are popular models with widespread adoption 
\cite{hinton_2012, dahl_2012,bahdanau2014neural,cho2014learning, kalchbrenner2013recurrent, 
oriol_neural_2015, antol_iccv15, ren_nips15}, 
Bidirectional RNNs have been utilized relatively infrequently \cite{huang_arxiv15,sak2015fast,wang_arxiv15} 
and even more rarely as decoders \cite{mathias_nips15} -- we argue due to the lack of efficient inference approaches for these models. 

Wang \etal \cite{wang_arxiv15} used Bidirectional RNNs for image caption generation, 
but do not perform bidirectional inference, rather simply use Bidirectional RNNs to rescore candidates. Specifically, 
at inference time they decompose a Bidirectional RNN into two independent Unidirectional RNNs, apply standard Beam Search in each direction, 
and then rerank these two collection of beams 
based on the max probability of each beam under the forward or backward model. 
We compare to this method and show that approximate joint optimization via Bidirectional Beam Search 
leads to better performance in the fill-in-the-blank image captioning task. 

Most related to our work is that of Burglund \etal \cite{mathias_nips15}, which studies generating missing sections of sequential data 
in an unsupervised setting using Bidirectional RNNs. 
They propose three probabilistically justified approaches to fill these gaps by drawing samples from the full joint. 

Their first model, Generative Stochastic Networks (GSN), resamples the output $\yb_t$ at a random time $t$ from the conditional output 
$\pr(\yb_t \mid Y_{[1:T] \setminus t})$. For a blank of length $w$, resampling each output tokens $M$ times requires $wMT$ passes of the RNN. 
Thus, the cost of producing a sample with the GSN method scales linearly with the size of the gap and requires a full pass of the Bidirectional RNN.
Their second approach, NADE, trains a model specifically for filling in the blank -- 
\ie at train time, some inputs are set to a specific `missing' token to indicate the content that needs to be generated. 
At inference time, the inputs from the gap are set to this token and sampled from the resulting conditional. 
Note that this approach is `trained to fill in gaps' and as such requires training data of this kind. To contrast, this is a new model for filling in gaps, while 
we propose a new inference algorithm, which can be broadly applied to any generative bidirectional model. 
Finally, they propose a third sampling approach 
based on a Unidirectional RNN which draws from the conditional $\pr(\yb_t \mid Y_{[1:T] \setminus t})$; 
however, as the model is a left-to-right Unidirectional RNN, this term requires computing the likelihood of the remaining sequence given each 
possible token at time $t$. This costly approach requires $w|\calY|MT$ steps of the RNN and is intractable for large vocabularies.

\vspace{\sectionReduceTop}
\section{Preliminaries: RNNs and Beam Search}
\label{sec:prelim}
\vspace{\sectionReduceBot}

We begin by establishing notation and reviewing RNNs and standard Beam Search for completeness. While our exposition 
details the classical RNN updates, the techniques developed 
in this paper are broadly applicable to any recurrent neural architecture (\eg LSTMs~\cite{hochreiter_nc97} or GRUs~\cite{cho_ssst14}). 

\textbf{Notation.} 
Let $X = (\xb_1, \xb_2,  \ldots, \xb_T)$ denote an input sequence, where $\xb_t$ is an input vector at time $t$. 
Similarly, let $Y = (\yb_1, \yb_2, \ldots, \yb_T)$ denote an output sequence, where $\yb_t$ is an output vector at time $t$. 
To avoid notational clutter, our exposition uses the same length for both input and output sequences ($T$); however, this is not a restriction in theory or practice. 
Given integers $a,b$, we use the notation $Y_{[a:b]}$ to denote the sub-sequence $(\yb_a, \yb_{a+1}, \ldots, \yb_b)$; thus, $Y = Y_{[1:T]}$  by convention. 
Given discrete variables $Y$, we generalize the classical maximization notation $\argmax_{Y \in \calY} \,\, f(Y)$ to find the 
(unique) top $B$ states with highest $f(Y)$ via the notation $\topB_{Y \in \calY} \,\, f(Y)$. 

\begin{figure*}[t]
\centering
\subfloat[Unidirectional RNN]{\includegraphics[trim=0pt 0pt 0pt 0pt, clip=true,width=0.28\textwidth]{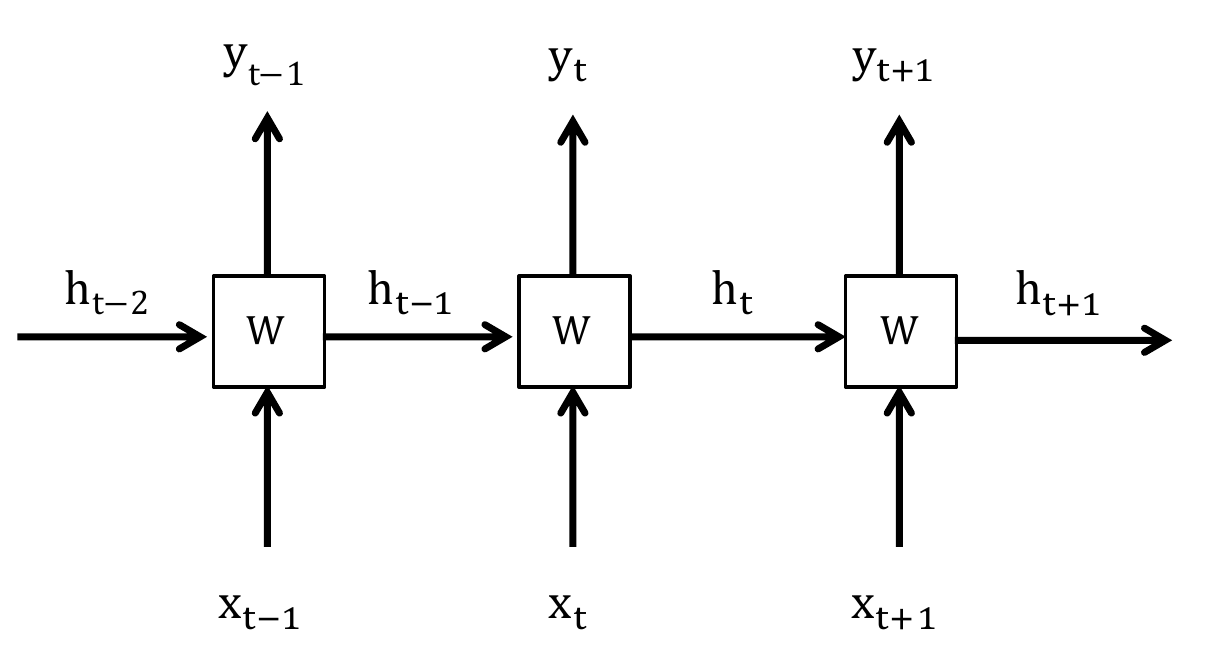}\label{fig:urnn}}
\hspace{10px}
\subfloat[Bidirectional RNN]{\includegraphics[trim=0pt 0pt 0pt 0pt, clip=true,width=0.28\textwidth]{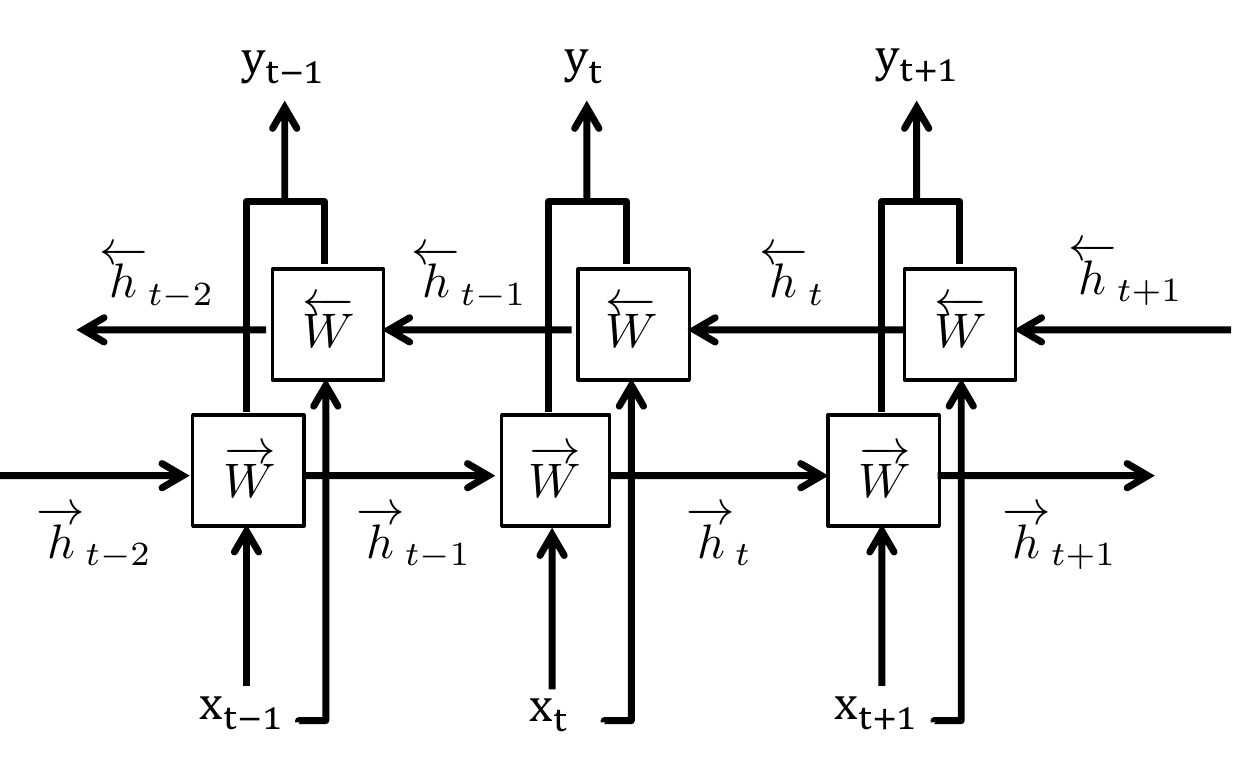}\label{fig:birnn}}
\hspace{10px}
\subfloat[Example Run of Beam Search]{\includegraphics[trim=0pt 0pt 0pt 0pt, clip=true, width=0.32\textwidth]{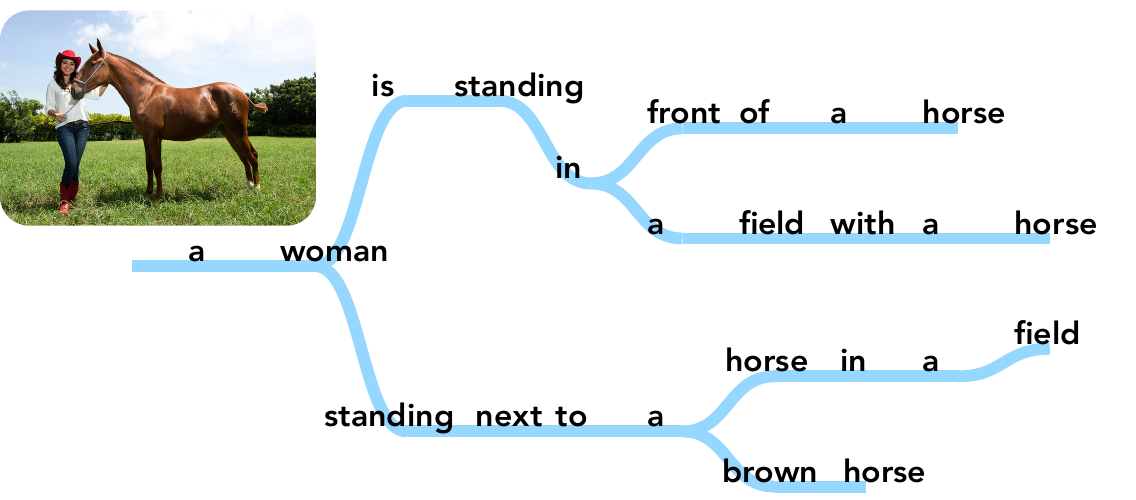}\label{fig:bs}}\\[10pt]
\caption{\textbf{Different architectures of RNNs and left-to-right Beam Search.} 
(a) The prediction of variable $\yb_t$ only depends on the \emph{past} in  URNNs. 
BiRNNs (b) can consider both \emph{past} and \emph{future}. 
(c) shows the search tree for beam search in a URNN with a beam width of $B{=}4$. }
\vspace{-12pt}
\label{fig:prelim}
\end{figure*}

\textbf{\urnn (URNNs)} model the probability of $\yb_t$ given the history of inputs $\xb_1, \ldots, \xb_{t}$
by ``compressing'' the history into a hidden state vector $\hb_t$ such that
\begin{subequations}
\begin{align}
\centering
\pr(\yb_t \mid X_{[1:t]}) &= \phi(W_y \hb_{t} + \bb_y) \label{eqn:urnnpr}\\
\hb_t 			&= \tanh(W_x \xb_t + W_h \hb_{t-1} + \bb_h)
\end{align}
\label{eq:urnn}
\end{subequations}\\[-10pt]
\noindent where $W_x, W_h, W_y, \bb_h,$ and $\bb_y$ are learned parameters defining the transforms from the input 
$\xb_t$ and hidden state $\hb_{t-1}$ to the output $\yb_t$ and updated hidden state $\hb_t$.
In applications with symbol sequences as output (such as image captioning), 
the nonlinear function $\phi$ is typically the softmax function which produces a distribution over the output 
vocabulary $\calY$. An example left-to-right Unidirectional RNN architecture is shown in Figure \ref{fig:urnn}.

\textbf{\brnns (BiRNNs)} (shown in Figure \ref{fig:birnn})
model both forward (positive time) and backward (negative time) dependencies 
via two hidden state vectors -- forward $\overrightarrow{\hb}_{t}$ and backward $\overleftarrow{\hb}_{t}$ 
-- each with its own update dynamics and corresponding weights. For a BiRNN, we can write
the probability of the token $\yb_t$ given the input sequence as

\vspace{\subsectionReduceTop}
\begin{subequations}
\vspace{\subsectionReduceBot}

\begin{align}
\centering
\pr(\yb_t \mid X_{[1:T]}) &= \phi(\underbrace{\overrightarrow{W}_y \overrightarrow{\hb}_{t}}_{\text{Forward score}} + \underbrace{\overleftarrow{W}_y \overleftarrow{\hb}_{t}}_{\text{Backward score}} + \bb_y) \label{eqn:brnnpr}\\
           \quad \overrightarrow{\hb}_t & = \sigma(\overrightarrow{W}_x \xb_t + \overrightarrow{W}_h \overrightarrow{\hb}_{t-1} + \overrightarrow{\bb}_h) \\ 
                    \overleftarrow{\hb}_t & = \sigma(\overleftarrow{W}_x \xb_t + \overleftarrow{W}_h \overleftarrow{\hb}_{t+1} + \overleftarrow{\bb}_h) \hspace{3pt}
\end{align}
\label{eq:brnn}
\end{subequations}\\[-20pt]

\noindent\textbf{BiRNNs as URNNs.}
Consider a Bidirectional RNN with the output nonlinearity $\phi$ defined as the softmax function $p_i = \phi_i(\sb) = e^{s_i} / \sum _j e^{s_j}$ 
It is straightforward to show that the conditional probability of $\yb_t$ given all other tokens can be written as 
\begin{eqnarray*}
\pr(\yb_t \mid X_{[1:T]})&=&\phi(\overrightarrow{W}_y \overrightarrow{\hb}_{t} + \overleftarrow{W}_y \overleftarrow{\hb}_{t} + \bb_y) \\
																			&\propto& \phi\left(\overrightarrow{W}_y \overrightarrow{\hb}_{t} + \tfrac{\bb_y}{2}\right) \phi\left(\overleftarrow{W}_y \overleftarrow{\hb}_{t} + \tfrac{\bb_y}{2}\right)
\end{eqnarray*} 
where the resulting terms in the proportionality resemble the URNNs output equation in Eq.~\ref{eqn:urnnpr}.
Intuitively, this expression shows that the output of a Bidirectional RNN with a softmax output layer can be equivalently expressed as the 

product of the output from two independent but oppositely directed URNNs with specifically constructed weights, renormalized after multiplication. 
This construction also works in reverse such that an equivalent Bidirectional RNN can be constructed from two independently trained but oppositely directed  URNNs. 
As such, we will consider a Bidirectional RNN as consisting of a forward-time model $\overrightarrow{\mbox{URNN}}$ and a backward-time model $\overleftarrow{\mbox{URNN}}$.

\textbf{RNNs for decoding} are trained to produce sequences conditioned on some encoded representation $X$. For machine translation tasks, $X$ may represent an encoding of some source language sequence to be translated and $Y$ is the translation. 
For image captioning, $X$ is typically a dense vector embedding of the image produced by a Convolutional Neural Network (CNN)~\cite{vgg_2014}, and $Y$ is a sequence of 1-hot encoding of the words of the corresponding image caption. 
Regardless of its source, this encoded representation is considered the first input $\xb_0$ and for all remaining time steps 
$\xb_t = \yb_{t-1}$, such that decoder RNNs are learning to model 
$\pr( \yb_t | \yb_{t-1}, ..., \yb_1, \xb_0)$. This is the setting of interest in this paper, but we drop this explicit dependence on the 
encoding $\xb_0$ to reduce notational clutter in later sections.

\textbf{Beam Search (BS).} 
Maximum a posteriori (MAP) (or more generally, M-Best-MAP \cite{batra_uai12, nilsson_sc88}) inference in RNNs consists of finding the most likely sequence under the model. 
The primary difficulty for decoding is that the number of possible $T$ length sequences grows exponentially as $|\calY|^T$, 
so approximate inference algorithms are employed. Due to this exponential output space and the dependence on previous outputs,   
exact inference is NP-hard in the general case. 
Beam Search (BS) is a greedy heuristic search algorithm that traverses the search tree using breadth-first search, while only expanding the most promising nodes at each depth.
Specifically, BS in \urnns involves maintaining and expanding the top-$B$ highest-scoring partial hypotheses, called \emph{beams}.  
Let $Y_{[1:t]} = (\yb_1, \ldots, \yb_t)$ denote a partial hypothesis (beam) at time $t$. 
We use the notation $\Yb_{[1:B], [1:t]} = (Y_{1, [1:t]}, Y_{2, [1:t]}, \ldots, Y_{B, [1:t]})$ to denote a collection of $B$ beams.
BS begins with empty beams, $Y_{b, 0} = (\yb_{b,0})$, where $\yb_{b,0} = \emptyset, \,\, \forall b$ and proceeds in a left-to-right manner up to time $T$ or until a special \texttt{END} token is generated. 
At each time $t$, BS considers the space of all possible beam extension $\calY_t = \Yb_{[1:B], [1:t-1]} \times \calY$ and selects the top-$B$ high-scoring $t$-length beams among this expanded hypothesis space. We can formalize this search for optimal updated beams $\Yb_{[1:B], [1:t]}$ as\\[-8pt]
\begin{align}
\topB_{\calY_t}~~\log\pr\left(Y_{[1:t]}\right) = \sum_{i=1}^t \log\pr(\yb_i \mid \yb_{i-1}, \ldots, \yb_{1}). \nonumber
\end{align}\\[-8pt]
\noindent Each log probability term in the above expression can be computed via a forward pass in \urnns such that implementing the $\topB$ operation simply requires sorting $B|\calY_t|$ values. An example run of BS on a left-to-right URNN is shown in Figure \ref{fig:bs}.

\vspace{\sectionReduceTop}
\section{Bidirectional Beam Search (BiBS)}
\label{sec:bsurnns}
\vspace{\sectionReduceBot}

\begin{figure*}[t]
\centering
\includegraphics[width=0.9\textwidth]{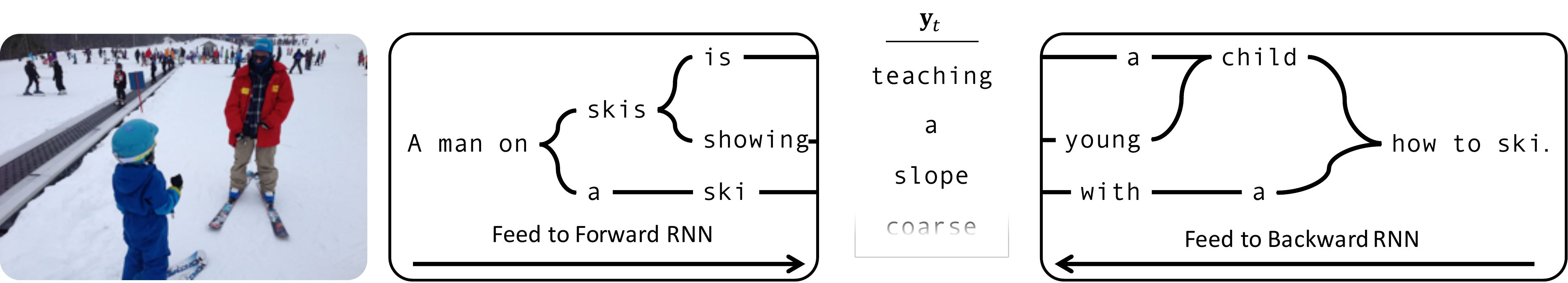}
\small
\begin{equation}
~~~~~~Y_{[1:B],[1:t]} = \topB_{b,\yb_t,b'} ~~~~~ \log\pr(Y_{b,[1:t-1]}) ~+~ \log\left(\pr\left(\yb_t | Y_{b,[1:t-1]}\right)\pr\left(\yb_t \mid Y_{b',[t+1:T]}\right)\right) ~+~ \log\pr(Y_{b',[t+1:T]}) \label{eq:update}
\end{equation}
\caption{\textbf{Overview of Bidirectional Beam Search (BiBS).} Starting from a set of $B$ complete sequences $Y_{[1,B],[1,T]}$, BiBS alternately performs left-to-right and right-to-left beam searches to greedily optimize an approximation of the probability of the entire sequence. In the example above, a left-to-right beam search is advancing the beams at time $t$ by considering all possible connections between the current left-to-right beams and the previous right-to-left beams through any token in the dictionary $\calY$. The terms in this joint approximation (written in  \eqnref{eq:update}) can be efficiently computed by the forward and backward Unidirectional RNNS and sorted to find the $\topB$ extensions.}
\label{fig:bibs}
\vspace{-10pt}
\end{figure*}

We begin by analyzing the decision made by left-to-right 
Beam Search at time $t$. 
Specifically, at each time $t$, 
we can factorize the joint probability $\pr(Y_{[1:T]})$ in a particular way: \\[-16pt]

{\small
\begin{equation}
\pr(Y_{[1,T]}) = 
\pr(Y_{[1,t-1]}) \pr(\yb_t | Y_{[1:t-1]}) \pr(Y_{[t+1:T]} \mid \yb_t, Y_{[1:t-1]})
\label{eq:joint}
\end{equation}
}\\[-6pt]
\noindent This left-to-right decomposition of the joint around $\yb_t$ is comprised of three terms
\begin{compactenum}[1)] 
\item the `marginal' of the sequence prior to $\yb_t$: $\pr(Y_{[1:t-1]})$, 
\item the conditional of $\yb_t$ given this past: $\pr(\yb_t | Y_{[1:t-1]})$, 
and 
\item  the conditional of the remaining sequence after $\yb_t$ given all prior terms: $\pr(Y_{[t+1:T]} \mid \yb_t, Y_{[1:t-1]})$. 
\end{compactenum}
If we consider choosing $\yb_t$ to maximize this joint, the first two terms can be computed exactly via the forward pass of a left-to-right URNN given the existing sequence; 
however, the third term cannot be exactly computed because it depends on all futures. Even approximating the third term with beams requires 
re-running beam search for each possible setting of $\yb_t$, which is  prohibitively expensive. 

One way of interpreting left-to-right BS is to view it as approximating the joint in \eqnref{eq:joint} with just the first two terms. 
Specifically, if we assume that $\pr(Y_{[t+1:T]} \mid \yb_t, Y_{[1:t-1]})$ is uniform, \ie all futures are equally likely given the sequence so far, 
then BS is picking the optimal $\yb_t$. 
This approximation does not hold in practice and results in poor performance for 
fill-in-the-blank tasks where all future sequences are not equally likely by design.  
In this section, we consider an alternative approximation and derive our BiBS approach.

\textbf{Efficiently Approximating the Future.} 
In order to derive a tractable approximation to this third term (and by proxy the full joint), we make two simplifying assumptions (which we know will be violated
in practice, but result in an efficient approximate inference algorithm). 
First, we assume that future sequence tokens are independent of past sequence tokens given $\yb_t$, \ie we treat RNNs as first-order Markov around $\yb_t$ during inference. 
Second, we assume that $\pr(\yb_t)$ is uniform, 
avoiding the need to estimate marginal distributions over $\calY$ for all time steps. Under these assumptions, we write the 
conditional probability of the remaining sequence tokens given the past sequence as 
\begin{eqnarray}
\pr(Y_{[t+1:T]} \mid Y_{[1:t]}) &=& \pr(Y_{[t+1:T]} \mid \yb_t) \label{eq:approx1} \nonumber \\
&\propto& \pr(\yb_t \mid Y_{[t+1:T]})\pr(Y_{[t+1:T]})~~~~~~~\label{eq:approx2}
\end{eqnarray}
Notice that the resulting terms are exactly the output of a right-to-left Unidirectional RNN. Substituting Eq.~\ref{eq:approx2} into Eq.~\ref{eq:joint}, we arrive at an expression that is proportional to the full joint, but comprised of terms which can be independently computed from a pair of oppositely-directed Unidirectional RNNs (or equivalently a Bidirectional RNN),
\begin{equation}
\underbrace{\pr(Y_{[1:t-1]}) \pr(\yb_t | Y_{[1:t-1]})}_{\mbox{Compute from }\overrightarrow{\mbox{\small URNN}}} \overbrace{\pr(\yb_t \mid Y_{[t+1:T]})\pr(Y_{[t+1:T]})}^{\mbox{Compute from }\overleftarrow{\mbox{\small URNN}}}~~
\label{eq:obj}
\end{equation}
Note that the two central conditional terms are proportional to the output of an equivalent softmax Bidirectional RNN as discussed in the previous section.

\textbf{Coordinate Descent.} Given some initial sequence $Y_{[1:t]}$, a simple coordinate descent algorithm could select a random time $t$ and update $\yb_t$ such that this approximate joint is maximized and repeat this until convergence. Computing Eq.~\ref{eq:obj} would require feeding $Y_{[1:t-1]}$ to the forward RNN and $Y_{[t+1:T]}$ to the backward RNN. Therefore, updating all outputs $M$ times in this approach would require $MT^2$ RNN steps (combined from both the forward and backward models). If we instead follow an alternating left-to-right then right-to-left update order, this can be reduced to $MT$ by reusing cached log probabilities from the previous direction. This algorithm resembles a beam search with $B=1$ which bases extensions on the value of Eq.~\ref{eq:obj}.

\newlength{\textfloatsepsave} \setlength{\textfloatsepsave}{\textfloatsep} 
\newlength{\floatsepsave} \setlength{\floatsepsave}{\floatsep} 
\setlength{\textfloatsep}{0.1cm}
\setlength{\floatsep}{0.1cm}
\begin{algorithm}
\small
\hrule\vspace{5pt}
\KwData{Given initial set of sequences $Y_{[1:B],[1:T]}$}
 $\overrightarrow{\theta}_{[1:B],[1:T]} = \overleftarrow{\theta}_{[1:B],[1:T]} = \boldsymbol{0}$\\
 \While{not converged}{
  \tcp{Update beams left-to-right}
  \For{$t=1,...,T$}{
    $\overrightarrow{\theta}_{[1:B],t}, \overrightarrow{h}_{[1:B],t} = \overrightarrow{\mbox{URNN}}\left(\overrightarrow{h}_{[1:B],t-1}, Y_{[1:B], t-1}\right)$\\
    $Y_{[1:B],t} = \topB \sum_{i=1}^{t} \overrightarrow{\theta}_{b,i}(\yb_{b,i}) + \sum_{j=t}^{T} \overleftarrow{\theta}_{b',j}(\yb_{b',j})$\\
  }
  \tcp{Update beams right-to-left}
  \For{$t=T,...,1$}{
    $\overleftarrow{\theta}_{[1:B],t}, \overleftarrow{h}_{[1:B],t} = \overleftarrow{\mbox{URNN}}(\overleftarrow{h}_{[1:B],t+1}, Y_{[1:B], t+1})$\\
    $Y_{[1:B],t} = \topB \sum_{i=t}^{T} \overleftarrow{\theta}_{b,i}(\yb_{b,i}) + \sum_{j=1}^{t} \overrightarrow{\theta}_{b',j}(\yb_{b',j})$\\
  }
 }
 \hrule\vspace{5pt}
 \caption{Bidirectional Beam Search (BiBS).}
\label{alg:bibs}
\end{algorithm}

\textbf{Bidirectional Beam Search.} Finally, we arrive at our full Bidirectional Beam Search (BiBS) algorithm by generalizing the simple algorithm outlined above to maintain multiple beams during each update pass. Given some set of initial sequences $Y_{[1:B],[1:T]}$ (perhaps from a left-to-right beams search), we alternate between forward (left-to-right) and backward (right-to-left) beam searches with respect to the approximate joint. We consider a pair of forward and backward updates a single round of BiBS.

Without loss of generality, we will describe a forward update pass of beam width $B$. At each time $t$, 
we have updated the first $t{-}1$ tokens of each beam such that we have partial forward sequences $Y_{[1:B],[1:t-1]}$ 
and the values $Y_{[1:B],[t+1:T]}$ have yet to be updated. 
To update the forward beams, we consider all possible connections between the current left-to-right beams and the 
right-to-left beams (held fixed from previous round) through any token in the dictionary $\calY$.
Our search space is then $\calY_t = Y_{[1:B],[1:t-1]} \times \calY \times Y_{[1:B],[t+1:T]}$ and 
$|\calY_t| = B \times |\calY| \times B$. 

Figure \ref{fig:bibs} shows an example left-to-right update step for image captioning as well as the precise update rule based on Eqn.~\ref{eq:obj} for this time step. For each combination of forward beam and backward beam, this objective can be computed easily from stored sum of log probabilities of each beam and conditional output of the forward and backward RNNs. Like standard Beam Search, the optimal extensions can be found exactly by sorting these values for all possible combinations. Our approach requires only $2BMT$ RNN steps to perform $M$ rounds of updates.  
Our approach is summarized in Alg.~\ref{alg:bibs} with $\overrightarrow{\theta}_{b,i}(\yb_{b,i})$ representing $\log\pr(\yb_{b,i} | Y_{b, [1:i-1]})$ from $\overrightarrow{URNN}$ and $\overleftarrow{\theta}_{b,i}(\yb_{b,i})$ representing $\log\pr(\yb_{b,i} | Y_{b, [i+1:T]})$ from $\overleftarrow{URNN}$.

\setlength{\textfloatsep}{\textfloatsepsave}
\setlength{\floatsep}{\floatsepsave}

\vspace{\sectionReduceTop}
\section{Experiments}
\label{sec:exp}
\vspace{\sectionReduceBot}

\begin{figure*}[t]
\centering 
\resizebox{\textwidth}{!}{
\small \def\arraystretch{1.1}
\begin{tabular}{c l @{\hspace{20pt}} c l}
\multirow{4}{*}{\includegraphics[height=1.1in]{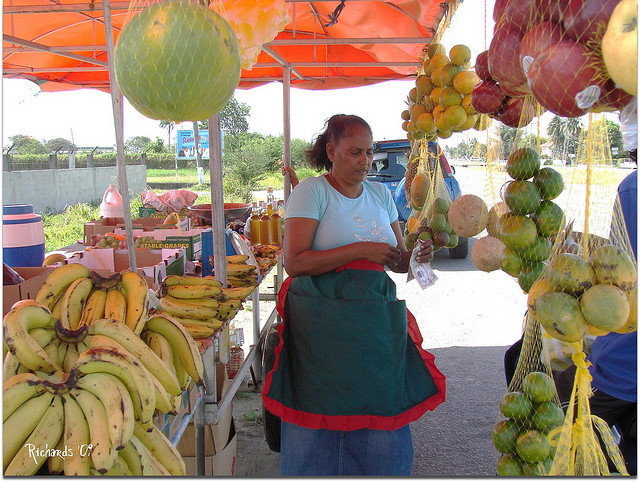}} &   &  \multirow{4}{*}{\includegraphics[height=1.1in,clip=true, trim=70px 0px 0px 0px]{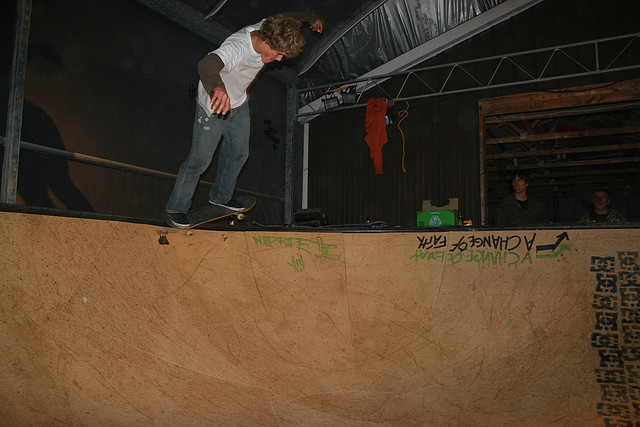}} &   \\
& \textbf{a) The woman has \underline{\color{purple} many bananas and other fruit} at her stand} && \textbf{a) A man \underline{\color{purple} is skateboarding on a ramp} in a basement} \\
& \textbf{b) The woman has \underline{\color{orange} a bunch of bananas on} at her stand} && \textbf{b) A man \underline{\color{orange} riding a skateboard up the} in a basement}\\
& \textbf{c) The woman has \underline{\color{red} holding a bunch of bananas} at her stand} && \textbf{c) A man \underline{\color{red} a trick on a skateboard} in a basement}\\
& \textbf{d) The woman has \underline{\color{blue} a large bunch of bananas} at her stand} && \textbf{d) A man \underline{\color{blue} doing tricks on a skateboard} in a basement}\\[25pt]
\multirow{4}{*}{\includegraphics[height=1.1in,clip=true, trim=100px 0px 0px 0px]{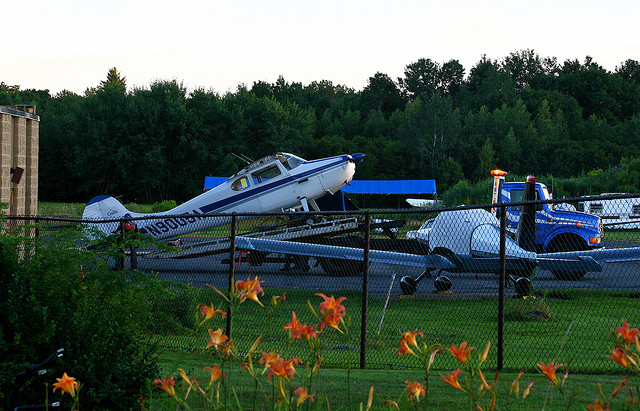}} &   &  \multirow{4}{*}{\includegraphics[height=1.1in]{figs/qual/img56.jpg}} &   \\
& \textbf{a) A number \underline{\color{purple} of small planes behind} a fence} && \textbf{a) A black \underline{\color{purple} and yellow bird with} a colorful beak} \\
& \textbf{b) A number \underline{\color{orange} of small planes on}  a fence} && \textbf{b) A black \underline{\color{orange} and yellow bird sitting} a colorful beak}\\
& \textbf{c) A number \underline{\color{red}  plane is parked near}  a fence} && \textbf{c) A black \underline{\color{red} a yellow bird with} a colorful beak}\\
& \textbf{d) A number \underline{\color{blue} of planes parked near}  a fence} && \textbf{d) A black \underline{\color{blue} and yellow bird with} in a basement}\\[25pt]
\multirow{4}{*}{\includegraphics[height=1.1in,clip=true, trim=100px 0px 0px 0px]{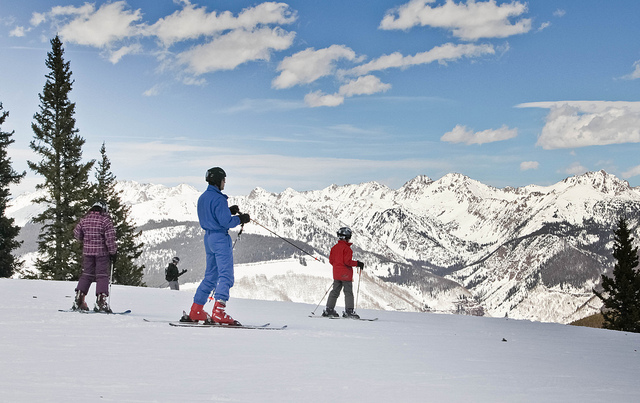}} &   &  \multirow{4}{*}{\includegraphics[height=1.1in, clip=true, trim=150px 0px 0px 0px]{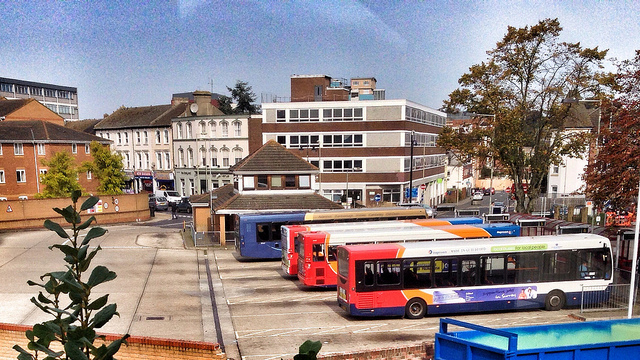}} &   \\
& \textbf{a) A group of \underline{\color{purple} people standing on top of a} snow covered slope} && \textbf{a) A row of \underline{\color{purple} transit buses sitting in} a parking lot} \\
& \textbf{b) A group of \underline{\color{orange} people on skis on a snowy} snow covered slope} && \textbf{b) A row of \underline{\color{orange} buses parked in a} a parking lot}\\
& \textbf{c) A group of \underline{\color{red} riding skis on top of a} snow covered slope} && \textbf{c) A row of \underline{\color{red} double decker buses parked} a parking lot}\\
& \textbf{d) A group of \underline{\color{blue} people standing on top of a} snow covered slope} && \textbf{d) A row of \underline{\color{blue} red buses parked in} a parking lot}\\[25pt]
\multirow{4}{*}{\includegraphics[height=1.1in,clip=true, trim=10px 0px 0px 0px]{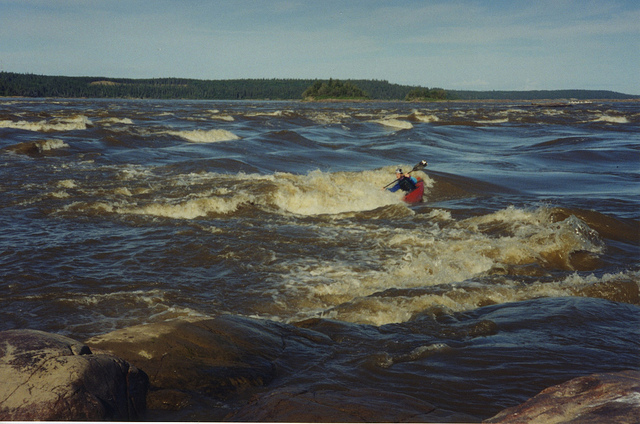}} &   &  \multirow{4}{*}{\includegraphics[height=1.1in, clip=true, trim=100px 0px 0px 0px]{figs/qual/img4885.jpg}} &   \\
& \textbf{a) The person is \underline{\color{purple} riding the waves} in the water} && \textbf{a) Two people \underline{\color{purple} riding a motorcycle to} the beach} \\
& \textbf{b) The person is \underline{\color{orange} is riding a wave} in the water} && \textbf{b) Two people \underline{\color{orange} on a motorcycle on} the beach}\\
& \textbf{c) The person is \underline{\color{red} person on a surfboard} in the water} && \textbf{c) Two people \underline{\color{red} riding a motorcycle on} the beach}\\
& \textbf{d) The person is \underline{\color{blue} is on a surfboard in} in the water} && \textbf{d) Two people \underline{\color{blue} on a motorcycle on} the beach}\\[25pt]
\end{tabular}
}\\[-8pt]
\resizebox{1\textwidth}{!}{
\setlength{\tabcolsep}{16pt}
\begin{tabular}{c c c c}
\tiny \textbf{\color{purple} a) Ground Truth} & \tiny \textbf{\color{orange} b) URNN-f} & \tiny \textbf{\color{red} c) URNN-b} & \tiny \textbf{\color{blue} d) BiRNN-BiBS}\\
\end{tabular}}\\[-2pt]
\caption{\textbf{Example fill-in-the-blank image caption completions generated by BS and BiBS.}   URNNs decoded with BS often produce blank reconstructions that clash with the remaining context on either side of the blank, while BiBS handles these transitions seamlessly.}
\label{fig:quals}
\end{figure*}


In this section, we evaluate the effectiveness of our proposed Bidirectional Beam Search (BiBS) algorithm for inference in BiRNNs. 
To examine the performance of bidirectional inference, we evaluate on tasks that require the generated sequence to fit well with existing structures 
both in the past and the future. We choose fill-in-the-blank style tasks where a number of tokens have been removed from a sequence and must be reconstructed. 
Specifically, we evaluate on fill-in-the-blank tasks on image captioning for the Common Objects in Context (COCO) \cite{coco} dataset and descriptions from the Visual Madlibs \cite{VisualMadlibs} dataset. 

\textbf{Baselines.} 
We compare our approach, which we denote \textbf{\texttt{BiRNN-BiBS}}, against several baseline approaches: 
\begin{compactitem}[\hspace{0pt}-]
\item \textbf{\texttt{URNN-f:}} that runs BS on a forward LSTM to produce B output beams (ranked by their probabilities under the forward LSTM), 
\item \textbf{\texttt{URNN-b:}} that runs BS on a backward LSTM to produce B output beams (ranked by their probabilities under the backward LSTM), 
\item \textbf{\texttt{URNN-f+b:}} that runs BS on forward and backward LSTMs to produce $2B$ output beams (ranked by 
the maximum of the probabilities assigned by the forward and backward LSTMs). The method used by Wang \etal \cite{wang_arxiv15}. 
\item \textbf{\texttt{BiRNN-f+b:}} that runs BS on two LSTMs (forward and backward) to produce $2B$ output beams (ranked by the sum of the log probabilities assigned by the forward and backward LSTMs). This lacks formal justification but we find it to be a reasonable heuristic for this task. 
\item \textbf{\texttt{GSN (Ordered):}} that samples tokens from the BiRNN for each time step. We found randomly selecting the time step as in \cite{mathias_nips15} resulted in poor performance on our tasks and instead perform updates in an alternating left-to-right / right-to-left order. For fairness, we compare at the same number of updates as our method and all sample sequences are reranked based on log probability. 
\end{compactitem}
All baselines perform inference on the same trained model that we train using \emph{neuraltalk2} \cite{karpathy_cvpr15} with standard maximum-likelihood training over complete human captions. 

\textbf{Evaluation.} For all models, we evaluate only the top beam from the sorted list returned by the algorithm. 
We compare methods on standard sentence-level metrics --  CIDEr\cite{vedantam_cvpr15}, Meteor\cite{banerjee2005meteor}, 
and Bleu\cite{papineni_acl02}  -- computed between the ground truth captions and the (full) reconstructed sentences. We note that 
the metrics are computed over the entire sentence (and not just the blank region) in order to capture the quality of the alignment 
of the generated text with the existing sentence structure. 
As a side effect, the absolute magnitude of these metrics are inflated due to the 
correctness of the context words, so we focus on the relative performance.  

\begin{table*}  \tiny 
\centering\def\arraystretch{1.025}
\setlength{\tabcolsep}{6pt}
\begin{center}
\resizebox{ 0.96 \textwidth}{!}{
\begin{tabular}{c @{} l  c  c  c  c  c  c  c c c c c c c@{}}
\toprule
& & & \multicolumn{3}{c}{$r$=25\%} & \multicolumn{3}{c}{$r$=50\%} & \multicolumn{3}{c}{$r$=75\%} \\
 \cmidrule(lr){4-6}
  \cmidrule(lr){7-9}
  \cmidrule(lr){10-12}
& &  & CIDEr& Bleu-4 & Meteor  & CIDEr& Bleu-4 & Meteor   & CIDEr& Bleu-4 & Meteor  \\
\midrule
  \multirow{6}{*}{\rotatebox{90}{Known Length~~}}
  & & URNN-f                             & 6.54  & 0.661 & 0.488 & 3.744 & 0.345 & 0.350 & 1.927 & 0.143 & 0.238 \\
  & & URNN-b                             & 6.58  & 0.668 & 0.491 & 3.931 & 0.372 & 0.356 & 2.476 & 0.219 & 0.259 \\
  & & URNN-f+b\cite{wang_arxiv15}        & 6.98  & 0.709 & 0.510 & 4.15  & 0.398 & 0.367 & 2.40 & 0.209  & 0.257 \\
  & & BiRNN-f+b                          & 6.94  & 0.705 & 0.508 & 3.99  & 0.385 & 0.361 & 2.24 & 0.201  & 0.252 \\
  & & GSN\cite{mathias_nips15} (Ordered) & 6.90  & 0.701 & 0.507 & 3.63  & 0.337 & 0.334 & 1.876 & 0.135 & 0.232 \\ 
  & & BiRNN-BiBS (ours)                  & \textbf{7.12} & \textbf{0.720} & \textbf{0.517} & \textbf{4.26} & \textbf{0.408} & \textbf{0.368}   & \textbf{2.57} & \textbf{0.228} & \textbf{0.265} \\
\midrule
  \multirow{6}{*}{\rotatebox{90}{Unknown Length~}}
  & & URNN-f                             & 5.607 & 0.569 & 0.440 & 4.232 & 0.432 & 0.370 & 2.594 & 0.268 & 0.269 \\
  & & URNN-b                             & 5.514 & 0.561 & 0.436 & 4.151 & 0.424 & 0.367 & 2.909 & 0.303 & 0.285 \\
  & & URNN-f+b\cite{wang_arxiv15}        & 5.632 & 0.570 & 0.440 & 4.377 & 0.451 & 0.376 & 2.924 & \textbf{0.306} & 0.287 \\ 
  & & BiRNN-f+b                          & 5.640 & 0.588 & 0.452 & 4.380 & 0.453 & 0.378 & 2.930 & 0.305 & 0.303 \\  
  & & GSN\cite{mathias_nips15} (Ordered) & 5.725 & 0.589 & 0.447 & 3.591 & 0.413 & 0.357 & 2.456 & 0.257 & 0.261 \\  
  & & BiRNN-BiBS (ours)                  & \textbf{5.935} & \textbf{0.614} & \textbf{0.460} & \textbf{4.40} & \textbf{0.454} & \textbf{0.380} & \textbf{2.936} & 0.305 & \textbf{0.288} \\
\bottomrule
\end{tabular}}
\vspace{5pt}
\caption{\textbf{Comparison of different approaches on Fill-in-the-Blank Image Captioning on COCO \cite{coco}.} $r$ is the percentage of removed words from sentence, $B$=5 by default.
BiBS consistently outperforms the baselines methods.}
\label{tab:coco}
\end{center}
\vspace{-20pt}
\end{table*}

\vspace{\subsectionReduceTop}
\subsection{Fill-in-the-Blank Image Captioning on COCO} 
\vspace{\subsectionReduceBot}
The COCO \cite{coco} dataset contains over 120,000 images, each with a rich set of annotations. 
This include five captions describing the content of each image, collected from Amazon Mechanical Turk workers. 
We propose a novel fill-in-the-blank image captioning task based on this data. 
Given an image $I$ and a corresponding ground truth caption $\yb_1, ..., \yb_T$ from the dataset, we remove a sequential portion of the caption such that we are left with a prefix $\yb_1, \ldots, \yb_s$ and suffix $\yb_e, \ldots, \yb_T$  consisting of the remaining words on either side of the blank. 
Using the image and the context from these remaining words, the goal is to generate the missing tokens $\yb_{s+1}, \ldots, \yb_{e-1}$. 
This is a challenging task that explores how well models and inference algorithms reason about the past and future during sequence generation. 
We first consider the known blank length setting (where the inference algorithm knows the blank length) and then generalize to the unknown blank length setting.

\textbf{Known Blank Length.}
In this experiment, we remove $r=$ 25\%, 50\%, or 75\% of the words from the middle of a caption for each image and task the model with generating the lost content. 
For example, at $r=$ 50\% the caption \emph{``A close up of flowers and plants inside of a bowl''} would appear to the system as the blanked caption \emph{``A close \_\_ \_\_ \_\_\_\_\_\_ \_\_\_ \_\_\_\_\_\_ \_\_\_\_\_\_ of a bowl''} and the generation task would then be to reproduce the removed subsequence of words \emph{``up of flowers and plants inside.''} 

As we are interested in bidirectional \emph{inference} (not learning), we train our models on the original COCO image captioning task (\ie we do not explicitly train to fill blanked captions). 
Like \cite{karpathy_cvpr15}, we use 5000 images for test, 5000 images for validation, and the rest for training. We evaluate on a single random caption per image in the test set.

The upper half of Table~\ref{tab:coco} reports the performance of our approach (BiBS) on this fill-in-the-blank inference task for differently sized blanks. We run GSN and BiBS for four full forward / backward passes of updates. Generally we find that bidirectional methods outperform unidirectional ones on this task. We find that BiBS outperforms all baselines on all metrics. We note that the nearest baselines in performance (\texttt{URNN-f+b}, \texttt{BiRNN-f+b}) are reranked from 2B beams. While BiBS operates in an alternating left-to-right and right-to-left fashion, it only ever maintains $B$ beams. 

Interestingly, the backward time \texttt{URNN-b} model consistently outperforms the forward time model \texttt{URNN-f} on all metrics and across all sizes of blanks. This may be due to the way the dataset was collected. When tasked with describing the content of an image, people often begin by grounding their sentences with respect to specific entities visible in the image (especially when humans are depicted). Given this, we would expect many more sentences to begin with the similar words such that generating the beginning of a sentence from the end would be an easier task. 

\figref{fig:quals} shows several qualitative examples, comparing completed captions from \texttt{URNN-f}, \texttt{URNN-b}, and our \texttt{BiRNN-BiBS} method with ground truth human annotations.
The unidirectional models running standard BS typically generate sentences that abruptly clash with existing words at the edge of the blank. 
For example in the top-left instance, the forward model produces the grammatically incorrect phrase ``bananas on at her stand'' and similarly the backward model outputs ``The woman has holding a bunch''.
This behavior is a natural consequence of the inability for these models to efficiently reason about the past and future simultaneously. While these unidirectional models struggle to reason about word transitions on either side of the blank, our BiRNN based BiBS algorithm typically produces reconstructions that smoothly fit with the context, producing a reasonable sentence ``The woman has a large bunch of bananas at her stand.''

\begin{figure*}[t]
\centering 
\resizebox{\textwidth}{!}{
 \large \def\arraystretch{1.5}
\begin{tabular}{l l l l l l}
\multirow{4}{*}{\includegraphics[height=1.2in, width=1.2in]{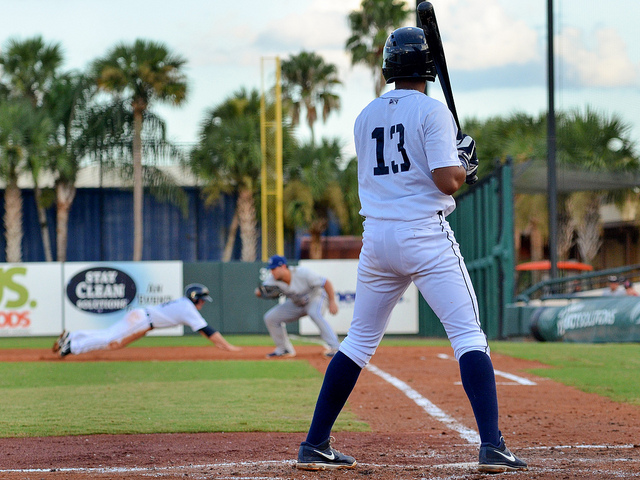}} & \textbf{(GT) Some baseball \underline{\color{purple} players are playing} a game}   
\hspace{10pt}\multirow{4}{*}{\includegraphics[height=1.2in,width=1.2in,clip=true, trim=100px 0px 0px 0px]{figs/qual/img864.jpg}} & \textbf{(GT) a woman in \underline{\color{purple} a red shirt is holding a} blue and orange kite}
\hspace{10pt}\multirow{4}{*}{\includegraphics[height=1.2in,width=1.2in,clip=true, trim=100px 0px 0px 0px]{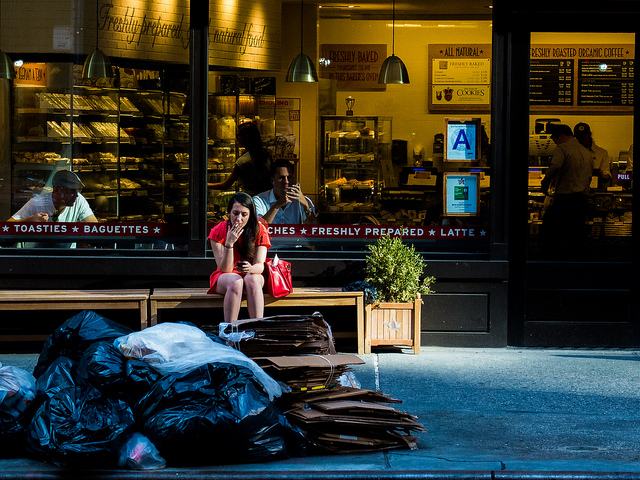}} & \textbf{(GT) this woman is \underline{\color{purple} sitting in front of a restaurant} smoking a cigarette} \\

& \textbf{(Init) Some baseball \underline{\color{orange} baseball bat during} a game}
& \textbf{(Init) a woman in \underline{\color{orange} a man is holding a} blue and orange kite} 
& \textbf{(Init) this woman is \underline{\color{orange} is working on a laptop and} smoking a cigarette}\\

& \textbf{(1st) Some baseball \underline{\color{red} baseball players in} a game} 
& \textbf{(1st) a woman in \underline{\color{red} a a hat is holding a} blue and orange kite} 
& \textbf{(1st) this woman is \underline{\color{red} sitting down on a laptop while} smoking a cigarette} \\

& \textbf{(2ed) Some baseball \underline{\color{blue} players playing in} a game} 
& \textbf{(2ed) a woman in \underline{\color{blue} a red shirt is holding a} blue and orange kite} 
& \textbf{(2ed) this woman is \underline{\color{blue} sitting down in a chair and} smoking a cigarette}

\end{tabular}
}\\
\vspace{10pt}
\caption{\textbf{Performance vs. Iteration.} Our model is initialized with right-to-left standard BS (Init) and updated alternatively from left-to-right (1st) and right-to-left (2nd).}
\label{fig:convergence_quals}
\vspace{-10pt}
\end{figure*}

This example also highlights a possible deficiency in our evaluation metrics; while a human observer can clearly tell which of the three sentences is most natural, the sentence level statistics of each are quite similar, with each sharing only the word banana with the ground truth caption ``The woman has many bananas and other fruit at her stand.'' Evaluating generated language is a difficult and open problem that is further complicated in the fill-in-the-blank task.

\textbf{BiBS Convergence.} To study the convergence of our approach, we consider the true joint probability of filled-in captions as a function of the number of rounds of BiBS. We compute the 
average of these joint log probabilities after each meta-iteration of BiBS, where we define a meta-iteration as a pair of full forward and backward update passes. We find that joint log 
probabilities drop quickly (reducing from -2.47 to -2.11 in a single meta-iteration), indicating high quality solutions are found from unidirectional initializations within only a few meta-iterations of BiBS. In practice, we find the beams typically converge in 1 to 2 meta-iterations for fill-in-the-blank image captioning. Figure \ref{fig:convergence_quals} shows how predicted sentence completions change over meta-iterations of BiBS for three example images.

\textbf{Unknown Length Blanks.} While our method is designed for known blank lengths, we can apply BiBS as a black box inference algorithm over a range of blank lengths and then rank these solutions. We calibrate what lengths to search over by first generating the top-1 left-to-right beam $Y^f$ by only conditioning on words on the left of the blank and the right-to-left top-1 beam $Y^b$ by only conditioning words on the right side of the blank. Then, we define the range of lengths of the blank as 
$\min\{len(Y^f),\,len(Y^b)\}$ to $\max\{len(Y^f), \,len(Y^b)\}$
where, $len(Y)$ is the length of beam $Y$. We perform inference at each length in this range and select the highest probability completion across all lengths. The lower half of Table~\ref{tab:coco} reports the results. We find that BiBS outperforms nearly all baselines on all metrics (narrowly being bested by \texttt{URNN-f+b} at $r=75\%$ Blue-4). 
All methods perform worse at the unknown length task than when blank length is know, due largely to the difficulty comparing sequences of differing lengths.

\vspace{\subsectionReduceTop}
\subsection{Visual Madlibs}
\vspace{\subsectionReduceBot}

\begin{table}[t] \small 
\setlength{\tabcolsep}{7pt}\def\arraystretch{1.025}
\begin{center}
\resizebox{0.94\columnwidth}{!}{
\begin{tabular}{c l  c  c  c c}
\toprule
& & \multicolumn{2}{c}{type 7} & \multicolumn{2}{c}{type 12} \\
 \cmidrule(lr){3-4}
  \cmidrule(lr){5-6}
& & Bleu-1 & Bleu-2 & Bleu-1 & Bleu-2 \\
\midrule
\multirow{8}{*}{\rotatebox{90}{Known Length}}&URNN-f & 0.313 & 0.138 & 0.275 & 0.160\\
&URNN-b & 0.460 & 0.284 & 0.346 & 0.213\\
&URNN-f+b\cite{wang_arxiv15} & 0.447 & 0.275 & 0.347 & 0.214\\
&BiRNN-f+b &0.448 & 0.275& 0.347& 0.213\\ 
&GSN\cite{mathias_nips15} (Ordered) &0.427 & 0.28& 0.148& 0.099\\ 
& BiRNN-BiBS (ours) & \textbf{0.470} & \textbf{0.389} & \textbf{0.353} & \textbf{0.216}\\
 \cmidrule{2-6}
& nCCA &  0.56 & 0.1 & 0.46 &0.07\\
& nCCA(box) & 0.60 & 0.11 & 0.48 & 0.08\\ 
\midrule
\multirow{8}{*}{\rotatebox{90}{~~~~~~~Unknown Length}}&URNN-f & 0.317 & 0.155 & 0.285 & 0.174\\
&URNN-b & 0.334 & 0.184 & 0.309 & 0.186\\
&URNN-f+b & 0.334 & 0.181 & 0.302 & 0.184\\
&BiRNN-f+b & 0.343 & 0.195  & 0.291  & 0.190\\ 
&GSN\cite{mathias_nips15} (Ordered) &0.348 & \textbf{0.203} & 0.270& 0.184\\ 
\cmidrule{2-6}
& BiRNN-BiBS & \textbf{0.351} & 0.197 & \textbf{0.31} & \textbf{0.190} \\
\bottomrule
\end{tabular}}
\vspace{3pt}
\caption{\textbf{Comparison of different approaches on the Visual Madlibs task} using BLEU-1 and BLEU-2. $B$= 5 by default. }
\vspace{\captionReduceBot}
\label{tab:madlib}
\end{center} 
\vspace{-10pt}
\end{table}

In this section, we evaluate our approach on the Visual Madlibs\cite{VisualMadlibs} fill-in-the-blank description generation task. 
The Visual Madlibs dataset contains 10,738 images with 12 types of fill-in-the-blank questions answered by 3 workers on Amazon Mechanical Turk.  
We use \emph{object's affordance} (type 7) and \emph{pair's relationship} (type 12) fill-in-the-blank questions as these types have blanks in the middle of questions. 
For instance, \emph{People could \underline{relax on} the couches.} and \emph{The person is \underline{putting food in} the bowl.} 
We use 2000 images for validation, train on the remaining images from the train set, and evaluate on their 2,160 image test set. 
To the best of our knowledge, ours is the first paper to explore the performance of CNN+LSTM text generation for this task. 

We compare to two additional baselines for these experiments, nCCA\cite{gong_2014} and the nCCA(box) method implemented in the Visual Madlibs paper \cite{VisualMadlibs}. 
nCCA maps image and text to a joint-embedding space and then finds the nearest neighbor from the training set to this embedded point. 
We note that this is a retrieval and not a description generation technique such that it cannot be directly compared with BiBS and report it only for the sake of completeness.
nCCA(box) operates similarly to nCCA, but extracts visual features from the ground-truth bounding box of the relevant person or object referred to in the question and thus is an `oracle' result that makes use of ground truth. 

We again use the \emph{neuraltalk2} \cite{karpathy_cvpr15} framework to train a CNN+LSTM model for both \emph{object's affordance} (type 7) and \emph{pair's relationship} (type 12) question types. 
We evaluate on the test data using Bleu-1 and Bleu-2 (to be consistent with \cite{VisualMadlibs}).
Table~\ref{tab:madlib} shows the results of this experiment for known blank length (top) and unknown blank length (bottom) settings. For known blank length, we find that BiBS outperforms the other generation based baselines in both question types and is competitive with the retrieval based nCCa techniques, greatly outperforming the nCCA retrieval and nCCA(box) oracle methods on Bleu-2. For unknown blank lengths, BIBS similarly outperforms the baselines, but is narrowly bested by GSN\cite{mathias_nips15} (Ordered) for type-7 questions under the Bleu-2 metric. We note that the
nCCA techniques have not been evaluated in this setting.

\vspace{\sectionReduceTop}
\section{Conclusions}
\label{sec:conclusions}
\vspace{\sectionReduceBot}
In summary, we presented the first approximate inference algorithm for 1-Best (and M-Best) decoding in bidirectional neural sequence models (RNNs, LSTMs, GRUs, \etc). 
We study our method in the context of a novel fill-in-the-blank image captioning task which evaluates how well sequence generation models and 
their associated inference algorithms incorporate known information from both the past and future to `fill in the gaps'. 
This is a challenging setting and we demonstrate that 
standard Beam Search is poorly suited for this task. We develop a Bidirectional Beam Search (BiBS) algorithm based 
on an approximation of the full joint distribution over output sequences 
that is efficient to compute in Bidirectional Recurrent Neural Network models. 
To the best of our knowledge, this is the first algorithm for top-$B$ MAP inference in Bidirectional RNNs. 
We have demonstrated that BiBS outperforms natural baselines at both fill-in-the-blank image captioning and Visual Madlibs. 
Future work involves generalizing these ideas to tree-structured or more general recursive neural networks~\cite{socher2011parsing}, and producing diverse M-Best sequences~\cite{batra_eccv12, vijay_dbs_2016}.

{\small
\textbf{Acknowledgements}
We thank Rama Vedantam for initial brainstorming.  
This work was funded in part by the following awards to DB: NSF CAREER, ONR YIP, ONR Grant N00014-14-1-0679, ARO YIP,
and NVIDIA GPU donations.
SL was supported in part by the Bradley Postdoctoral Fellowship.
Conclusions contained herein are the authors and are not to be interpreted as necessarily representing the
official policies or endorsements, either expressed or implied, of the U.S.~Government, or any sponsor.}

{
\bibliographystyle{style/ieee}
\bibliography{bib/strings,bib/dbatra}
 }

\end{document}